%% file: main.tex
\documentclass[sigconf]{aamas} 
\setcopyright{ifaamas}
\acmConference[AAMAS '21]{Proc.\@ of the 20th International Conference on Autonomous Agents and Multiagent Systems (AAMAS 2021)}{May 3--7, 2021}{Online}{U.~Endriss, A.~Now\'{e}, F.~Dignum, A.~Lomuscio (eds.)}
\copyrightyear{2021}
\acmYear{2021}
\acmDOI{}
\acmPrice{}
\acmISBN{}

\acmSubmissionID{XXX}

\usepackage{url}
\usepackage{graphicx}
\usepackage{amsmath}
\usepackage{booktabs}
\usepackage{amsthm}
\usepackage{amsfonts}       
\usepackage{caption}
\usepackage{subcaption}
\usepackage{paralist}   
\usepackage{hyperref}       
\usepackage{nicefrac}       
\usepackage{balance}

\DeclareMathOperator*{\argmax}{arg\,max}
\DeclareMathOperator*{\argmin}{arg\,min} 
\usepackage{array,booktabs,makecell}
\newtheoremstyle{own}%
    {3pt}
    {3pt}
    {}
    {}
    {\color{black}\bfseries}
    {:}

    {}

\theoremstyle{definition}

\theoremstyle{definition}
\newtheorem{definition}{Definition}

\renewcommand{\cite}[1]{\citep{#1}}

\newcommand{\real}{\ensuremath{\mathbb{R}}}

\title{Deceptive Reinforcement Learning for Privacy-Preserving Planning}

\author{Zhengshang Liu, Yue Yang, Tim Miller, and Peta Masters}
\affiliation{%
  \institution{School of Computing and Information Systems, The University of Melbourne}
  \institution{\{zhengshangl,yuey16\}@student.unimelb.edu.au,~\{peta.masters,tmiller\}@unimelb.edu.au}
  }

\input{0-abstract.tex}

\begin{document}

\pagestyle{fancy}
\fancyhead{}

\maketitle

\input{1-introduction.tex}
\input{2-related-work}
\input{3-model.tex}
\input{4-computational-evaluation.tex}
\input{5-human-evaluation.tex}

\input{6-conclusions.tex}

\bibliographystyle{ACM-Reference-Format}
\balance
\bibliography{references}

\newpage

\input{appendix}

\end{document}

%% file: 0-abstract.tex
\begin{abstract}
In this paper, we study the problem of deceptive reinforcement learning to preserve the privacy of a reward function. Reinforcement learning is the problem of finding a behaviour policy based on rewards received from exploratory behaviour. A key ingredient in reinforcement learning is a \emph{reward function}, which determines how much reward (negative or positive) is given and when. However, in some situations, we may want to keep a reward function private; that is, to make it difficult for an observer to determine the reward function used. We define the problem of privacy-preserving reinforcement learning, and present two models for solving it. These models are based on \emph{dissimulation} -- a form of deception that `hides the truth'. We evaluate our models both computationally and via human behavioural experiments. Results show that the resulting policies are indeed deceptive, and that participants can determine the true reward function less reliably than that of an honest agent.
\end{abstract}

\keywords{Reinforcement Learning; Deception; Dissimulation}

%% file: 1-introduction.tex
\section{Introduction}
\label{sec:introduction}

In this paper, we study the problem of deceptive reinforcement learning to preserve the privacy of reward functions. Reinforcement learning is a framework within which an agent learns a behaviour policy by interacting with its environment and responding to positive and negative rewards \cite{sutton2018reinforcement}. 
Within this framework, the \emph{reward function}, which determines when and how much reward (negative or positive) is given for each possible behaviour in a system, is critical. It defines the goals of the agent.

Situations frequently arise in which we do not want our goals to be known. Consider a military commander needing to conceal the purpose of troop movements; a crime-writer who must avoid giving away the end of the  story. In reinforcement learning, when we want to make it difficult for an observer to infer the final destination, we must prevent or delay them from determining the reward function used to learn a policy.

Deception involves fostering or maintaining false belief in the minds of others \cite{carson10}. \citeauthor{bb} defines two general types: \emph{dissimulation}, which `hides the truth' to avoid revealing information; and \emph{simulation}, which  `shows the false' enticing an observer to believe something that is not true. 
Several models of deceptive planning have been proposed in recent years \cite{masters2017deceptive,keren16,kulkarni18a,ornik2018deception}. However, these are model-based and require reasoning about the model structure to inform the dissimulation, so are not applicable to model-free MDPs.

In this paper, we define a more general model of dissimulation for preserving goal privacy. We present two methods: one based on ambiguity, in which the agent selects actions that maximise the entropy from the observer's point of view; and one based on \citet{masters2019goal}'s model for intention recognition using irrationality, which takes action selection as a weighted sum of honest and `irrational' behaviour. These methods use pre-trained Q-functions (or policies). Since Q-functions provide a measure of expected future reward for each action, they enable a general representation of the possibilities for action selection \cite{sutton2018reinforcement}.

\begin{figure}[!t]
\centering
\includegraphics[scale=0.55]{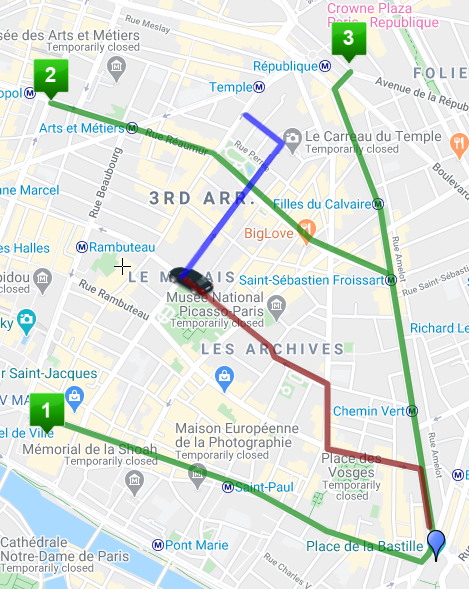}
\caption{Using dissimulation to deceive an observer about the final destination. Taking the red path, what is the final destination?}
\label{fig:deceptive-navigation-paris}
\end{figure}

Figure~\ref{fig:deceptive-navigation-paris} shows an example of dissimulation. An escort driver in Paris has three potential destinations (in green), starting from the blue point. The green routes are the optimal routes to each destination. If the driver takes the red route and is where the car is located, as the observer, what do you think the final destination is, knowing that the driver may be deceiving you? The path makes sub-optimal progress to all three destinations. If the driver turns right and follows the blue route, destination 1 is probably eliminated from the set of potential goals, but the blue path is valid for both destinations 2 and 3. Our ambiguity model would generate a path corresponding to the red path, and the red+blue path for destinations 2 and 3 once destination 1 is pruned as a possible path.

Such problems of deception are common and the use of AI for deception is gaining recent traction \cite{sarkadi2019modelling} in domains such as path planning \cite{masters2017deceptive}, military tactical planning \cite{ramirez2018integrated,santos2008deception}, countering cyber-attacks \cite{rowe2003counterplanning} and conjuring tricks  \cite{smith2016construction}.

We evaluate our models by using a na\"ive intention recognition system and via a human subject experiment with 69 non-na\"ive  participants. The intention recognition system and our participants were required to estimate the likelihood of different destinations in a path planning simulation. The results show that our agents are effective at hiding their reward compared to honest agents, but that, like honest agents, the true reward function becomes clearer as more actions are executed. The irrationality model deceives more than the ambiguity model, but receives less discounted expected reward on its real reward function.

%% file: 2-related-work.tex
 \section{Background and Related Work}
\label{sec:lit}
 
\subsection{Theory of Deception}

Deception, psychologists broadly agree, is a pejorative term for the fostering or maintenance of false belief in the minds of others \cite{carson10}. Computer science has necessarily widened the definition, first, to accommodate mindless machines incapable of belief (as such) and second, to allow for the emerging realisation (particularly from the field of social robotics) that deception is a fundamental aspect of intelligent behaviour, frequently beneficial not only to the deceiver but to the deceived \cite{shim13,wagner11}. 
Whereas deceptive AI has tended to focus on detection \cite{avrahami14}, ethical implications \cite{arkin12}, and the qualities that make a deceptive act most likely to succeed \cite{ettinger10}, military strategists Bell and Whaley \cite{whaley82,bb,bowyer82} provide a general theory focused on \emph{how} to deceive. Their non-judgemental definition is ``the distortion of perceived reality'' which they maintain can \emph{only} be achieved in one of two ways: by simulation (``showing the false'') or dissimulation (``hiding the true''). They propose three variations on each method and suggest that a deceptive strategy typically involves combinations of those six tactics in pursuit of some strategic objective. 
 
\subsection{Deceptive Planning}
In planning, deception is frequently associated with security and has become almost synonymous with privacy-protection \cite{chakraborti19}. Dissimulation in this context becomes the task of obscuring intent by maximising a plan's ambiguity \cite{kulkarni18,keren16}.
Obscuring intent assumes an observer engaged in intention recognition; and deceptive planning is commonly (though not exclusively)\footnote{See \cite{kulkarni18a} for an argument against.} conceived as an inversion of intention recognition \cite{dragan14,keren16,masters2017deceptive}.

In this paper, we invert a type of cost-based goal recognition \cite{ramirez10}. To generate a probability distribution over goals, they compare each goal's \emph{cost difference}, that is, the difference between the optimal cost of a plan via observed actions and the optimal cost of any alternative plan. The lower the cost difference, the higher the probability. 
\citet{vered16} take a similar approach but instead of cost difference use the ratio between the optimal cost of reaching each goal via the observations and the optimal cost per se. They propose two heuristics to minimise the computational effort in the context of online recognition, one of which suggests pruning a goal from consideration if observations deviate too far from the optimal behaviour.
\citet{masters2017deceptive} apply Bell and Whaley's theory to path-planning. They assume a na\"ive observer,  modelled as a probabilistic intention recognition system. The inputs are observations $\vec{o}$ and the output is a probability distribution across potential goals $P(G|\vec{o})$. 
An action is deceptive if, at that step, the probability of the real goal $g_r$ does not dominate the probability of some other goal: $P(g_r|\vec{o})\leq P(g)|\vec{o})$ for all $g \in G \setminus \{g_r\}$. They observe that \emph{every} path has one last deceptive point ($LDP$), even if it is the starting point, and show that there is a radius around goal within which trying to deceive is no longer valuable, and the agent should head directly to its true goal.
At the path level, they define deceptive density as inversely proportional to the number of truthful steps it contains; and deceptive extent by the distance remaining after the last deceptive point has been reached, that is, the optimal cost from $LDP$ to $g_r$.
\citet{kulkarni18} extend a similar approach to classical task planning, more general than path planning. 
Both approaches, however, are applicable only to model-based problems, so do not generalise to MDPs. 

\subsection{Deception in Markov Decision Processes}

\citet{ornik2018deception} present the comprehensive model of planning for deception in MDPs. Their model defines the notion of a \emph{belief-induced reward}, which is a reward that the agent receives, but that is also affected by the belief of an observer. This includes cases when the observer has only partial visibility of the environment. For example, the reward is received if the observer's belief is that the agent is not in the state that receives the reward, otherwise it receives some negative reward. \citeauthor{ornik2018deception} then show how to define optimal policies for belief-induced rewards, and present some examples of \emph{deceptive} belief-induced reward functions. However, their work is model-based, and further, they do not specify a dissimulative policy. 

\citet{karabag2019optimal} present a model-based solution to a different deceptive problem. In their problem, an agent is provided a policy to follow to achieve a goal, specified in linear temporal logic, but can instead follow a different deceptive policy, modelled from an MDP, to achieve the goal. The aim is to try to achieve both policies while minimising the likelihood of the supervisor knowing.

A closely related area of research is differential privacy for reinforcement learning \cite{vietri2020private,ma2019differentially,wang2019privacy}. The general approach to this is to modify Q-learning and policy-based reinforcement learning algorithms by e.g.\ adding Gaussian noise to the update rule \cite{wang2019privacy}. While the general idea is similar, there are a few major differences. First, our problem definition is motivated by \emph{strategic} deception based on theory of deception \cite{whaley82,bowyer82}, rather than on the idea of privacy per se. This difference manifests itself in the problem definition: we assume that reward functions are fully observable to the observer, but that the observer does not know which reward function the current policy is trained on. The work cited here,  on the other hand, assume that there is a single reward function and it is not observable. A privacy-preserving approach  like this is not strategic as it does not trade off against different goals. Further, it means that we explicitly measure the simulation of the policy, rather than the privacy of the reward function.  While we could frame our problem in a similar way to \citeauthor{wang2019privacy}, in strategic deception, it is uncommon for an observer not to have a model of likely goals for an actor. Second, we present a general model for MDPs, whereas \citet{wang2019privacy}, \citet{vietri2020private}, and \citet{ma2019differentially} are Q-learning and policy-gradient approaches. Finally, we measure the strategic deception achieve both in computational and human studies.

Some work in deceptive reinforcement learning investigates techniques to counter the deceptive strategies of other agents, such as the agent being fed incorrect reward signals \cite{huang2019deceptive}, deception in games and multi-agent systems \cite{banerjee2003countering,li2020effective,sakuma2008privacy}.

\subsection{Inverse Reinforcement Learning and Imitation Learning}

Our definition  of deceptive reinforcement learning is related to inverse reinforcement learning \cite{ng2000algorithms} and imitation learning \cite{ziebart2008maximum}. 
Inverse reinforcement learning is the problem of inferring a reward function given traces of an agent’s behaviour in a variety of circumstances and the sensory input to the agent. Imitation learning \cite{ziebart2008maximum} is similar to inverse reinforcement learning, but instead of inferring a reward function, the aim is to infer a policy.
These methods learn a reward function by observing e.g., a human complete the same task many times. The problem that we define in this paper could be framed as the problem of producing a policy that makes it difficult to perform inverse reinforcement or imitation learning. However, there are two key differences. First, in this paper, we aim to simply deceive for a single trace of behaviour, whereas these inverse learning problems require either a known optimal policy from which to generate traces, or a set of traces of behaviour. Despite this, there is clearly a related problem that is of interest in studying the problem of deception as obfuscating inverse reinforcement learning. Second, we define a set of possible reward functions, whereas inverse reinforcement learning starts with the set of all reward functions. The approach from \citet{wang2019privacy} above proposes a Q-learning-based solution for such a problem.  

%% file: 3-model.tex
\section{Models}
\label{sec:model}

In this section, we define privacy-preserving reinforcement learning and  present two solutions based on dissimulation.


\subsection{Problem Formalism}


\begin{definition}[Markov Decision Process (MDP) \cite{puterman2014markov}]
\label{defn:mdp}
An MDP is a tuple $\Pi = (S, A, T, r, \gamma)$, in which $S$ is a set of states, $A$ is a set of actions, $T(s,a,s')$ is a transition function from $S \times A \rightarrow 2^S$, which defines the probability of action $a$ going to state $s'$ from state $s$, $r(s,a,s')$ is the \emph{reward} received for the transition from executing action $a$ in state $s$ and ending up in state $s'$, and $\gamma$ is the discount factor.
The task is to synthesise a \emph{policy} $\pi : S \rightarrow A$ from states to actions that maximises expected reward over trajectories in $\pi$ for problem $\Pi$:
\[
\mathbb{E}[\sum_{t=0}^{T} \gamma^t R(s,\pi(s),s')]
\]
\end{definition}

A Q-function $Q : S \times A \rightarrow \real$ defines the value of selecting an action $a$ from state $s$ and then following the policy $\pi$, written $Q(s,a)$. An optimal policy $\pi$ can then be defined as $\pi(s) = \argmax_{a\in A} Q(s,a)$.

\begin{definition}[Belief-induced reward]
\citeauthor{ornik2018deception} \cite{ornik2018deception} define \emph{belief-induced} rewards to model rewards that are dependent on the reward function and the beliefs of an observer. Formally, this is a function $L : S  \times A \times S \times \mathcal{B}$, in which $\mathcal{B}$ is a set of beliefs. 
\end{definition}

\citeauthor{ornik2018deception} leave the actual instantiation of beliefs abstract, but the concept is that $L(s, a, s, B)$ is a reward that is some function of the belief of an observer and the real reward. 

Using a belief-inducted reward, the task of solving an MDP is to synthesise a \emph{policy} $\pi$ that maximises:
\begin{equation}
\label{eq:belief-inducted-expected-reward}
\mathbb{E}[\sum_{t=0}^{T} \gamma^t L(s, \pi(s), s', B)]
\end{equation}

To specify a deceptive reinforcement learning problem, we must instantiate $\mathcal{B}$ and define $L$. In our setting, $\mathcal{B} = \mathcal{R}$, as we aim to deceive about the particular reward function. For $L$, we need to define what is means to deceive about a reward function. 

First, we need to define the observer's task. This is an intention recognition task \cite{baker2009action} in which the observer derives a probability distribution over $\mathcal{R}$ that defines the probability $P(r_i \mid \vec{o_t})$ that the reward function $r_i$ is the true reward function, given $\vec{o_t}$, the sequence of observed state-action pairs up until time $t$. For example, the probability of the final destination of the each of the three locations outlined in Figure~\ref{fig:deceptive-navigation-paris}. Our deceptive models later present some ways to define this for an MDP.

\begin{definition}[Deceptive reinforcement learning for Reward-Function Privacy]
A \emph{deceptive reinforcement learning problem} is a tuple $\Pi = (S, A, T, r, \mathcal{R}, \gamma, L)$, in which $S$, $A$, $T$, $r$, and $\gamma$ are as in Definition~\ref{defn:mdp}, $\mathcal{R}$ is a set of possible reward functions such that $r \in \mathcal{R}$, which model the set of reward functions that an observer may believe are true, and $L$ is belief-inducted reward.
The task is to synthesise a \emph{policy} $\pi$ that maximises expected reward over trajectories in $\pi$ while also making it difficult for an observer to determine which reward function in $\mathcal{R}$ is the real reward function. 

Defining $L$ is not a straightforward task, and depends on the specific domain being used. Typically, it would be defined as some weighted measure of the reward and the level of deception, such as:
\begin{equation}
\label{eq:deceptive-L}
L(\vec{o_t}, \mathcal{R}) = (1-\omega) \cdot r(s,a,s') +  \omega \cdot d(\vec{o_t} \mathcal{R})
\end{equation}
%
%
%
in which $s$, $a$, and $s'$ are the state-action-state values of the last transition in $\vec{o_t}$; that is, the latest transition; $d(\vec{o_t}, \mathcal{R})$ is a measure of deception such as the simulation value defined by \citet{masters2017deceptive}, and $\omega \in [0,1]$ is a weighting factor for deception that determines how important the deception is. One difficulty in defining $\omega$ is that the rewards and the deception are of different magnitudes. Even if both are normalised, a  policy using dissimulation (hides the truth) may use subtle deception, meaning that any definition of $d(\vec{o_t},\mathcal{R})$ has to capture that subtly between honest and deceptive behaviour.
\end{definition}

The challenge of this problem is that it is difficult to model the intention recognition of the observer. For example, is the observer na\"{i}ve, in that they do not believe that they are being deceived? Or are they aware that they are being deceived? Or somewhere in the middle? If they have some awareness, what model of deception are they using in their own intention recognition model. For this reason, the straightforward model of just solving for Equation~\ref{eq:belief-inducted-expected-reward} is only optimal if our model of intention recognition is the same as the observers, which is unlikely.

In this paper, we present two 
solutions to this problem that do not have an explicit  model of an observer: an \emph{ambiguity model} and an \emph{irrationality model}. Instead, the two models use only the information available to them by their given policy.
We assume pre-trained Q-functions for all reward functions in $R_n$; or alternatively, pre-trained stochastic policies, but we only use Q-functions for the remainder of the paper. We use $Q_{r_i}$ to represent one trained on bogus reward function $r_i$. 

\subsection{Ambiguity Model}
In this model, an agent behaves ambiguously by selecting actions that have high Q-values not only for the real reward function but also for multiple bogus reward functions. As the trajectory progresses, fewer reward functions remain sensible, so these are pruned from consideration. Eventually, the policy selects actions only optimal for the true reward function. The final point before this occurs conforms to \citeauthor{masters2017deceptive}'s \emph{last deceptive point} \cite{masters2017deceptive}.

The main idea is for our policy to generate sequences of actions that have positive reward for several reward functions, including the true reward function. 
For this, we need a measure of how far a sequence of observed behaviour diverges from optimal behaviour. Observations $\Vec{o}$ consist of a sequence of tuples $(s, a)$. We measure how far observations diverge from the optimal solution by summing the Q-differences:
\begin{equation}
\label{eq:divergence-diff}
    \Delta_{r_i}(\Vec{o}) = \sum_{(s,a) \in \Vec{o}} \left(Q_{r_i}(s,a) - \max_{a' \in A} Q_{r_i}(s,a')\right)
\end{equation}
This formula is based on the definition by \citet{ramirez10}. If $\Vec{o}$ follows a sequence that is optimal for reward function $r_i$, then $\Delta_{r_i}(\Vec{o}) = 0$, and any sub-optimal behaviour has a negative divergence. Other definitions are possible, such as using cost ratio like \citet{vered16}; what matters is that $\Delta$ allows us to compare behaviour with respect to optimality. 

%
%

The probability that reward function $r_i$ is the true reward function $r$, from the perspective of an observer, is defined using a Boltzmann distribution:
\begin{equation}
    P(r_i \mid \Vec{o}) = \frac{exp\{ \Delta_{r_i}(\Vec{o})\}}{\sum_{r_j \in \mathcal{R}} exp\{ \Delta_{r_j}(\Vec{o})\}} \cdot P(r_i),
\end{equation}
in which $P(r_i)$ is the prior probability that $r_i$ is the true reward function, which can be uniform over $\mathcal{R}$ if this is unknown.
If $\Vec{o}$ is far from optimal for $r_i$ compared to other reward functions in $\mathcal{R}$, its probability will be lower relative to the other reward functions. This gives us a probability distribution over all reward functions in $\mathcal{R}$. As with $\Delta_{r_i}(\Vec{o})$, other models could be used to define this, but we use what is common in intention recognition models.

Our model uses this probability distribution to minimise information gain by the observer using Shannon entropy\footnote{Shannon entropy measures \emph{information gain}. Increasing uncertainty lowers information gain and increases entropy.} \cite{shannon1948mathematical} each time an action is chosen. 

We define the \emph{Q-gain} of action $a$ for reward function $r_i$ as: 
\[
G_{r_i}(s,a) = Q_{r_i}(s,a) - R_{r_i}(\Vec{o})
\]
in which $R_{r_i}(\Vec{o}) = Q_{r_i}(s', a') - Q_{r_i}(s_0, a_0)$ is the residual expected reward received so far in sequence $\Vec{o}$ where $(s',a')$ is the last pair in $\Vec{o}$ and $(s_0, a_0)$ is the first pair in $\Vec{O}$. Thus, $R_{r_i}(\Vec{o})$ represents the value of having arrived at state $s$ minus the reward of executing $a$, while $G_{r_i}(s,a)$ represents the gain that action $a$ gives compared to `remaining' in state $s$.
Intuitively, $G_{r_i}(s,a) < 0$ implies that action $a$ is moving `away' from the rewards given by $r_i$, and $G_{r_i}(s,a) > 0$ is moving `towards' the rewards.

Given a sequence of observations $\Vec{o}$, our model chooses the action that minimises the information gain for the observer:
\begin{equation}
\label{eq:entropy}
   \pi^D(\Vec{o}, s) = \argmin_{a \in A(s)} -\kappa \sum_{r_i \in \mathcal{R}} P(r_i \mid \Vec{o} \cdot a) \times log_2(P(r_i \mid \Vec{o} \cdot a))
\end{equation}
in which $A^+(s)$ is the set of actions with non-negative Q-gain for the real reward function $r$, and $\kappa$ is a normalising term.
Thus, an agent following policy $\pi^D$ will move ambiguously between all of the Q-functions to maximise entropy. Only evaluating actions in $A^+(s)$ ensure that progress is made towards the real goal.

However, sometimes a particular reward can become so irrational that it would be clear to an observer that this is no longer likely. We exclude such  reward functions from the entropy calculation by re-evaluating the bogus reward functions at each step of the plan, and excluding those would be irrational (negative Q-gain). This is similar to the pruning heuristic from \citet{vered2017heuristic}.

A reward function is pruned from the entropy calculation (set $\mathcal{R}$ in Equation~\ref{eq:entropy}) if $G_{r_i}(s,a) < \delta$, in which $\delta$ is a pruning parameter. If $\delta = 0$, a reward function is pruned because it offers no gain over the current state. If $\delta < 0$, the pruning would be less aggressive, allowing some actions that offer no gain. If $\delta = {-}\infty$, nothing would be pruned. At each step, all reward functions are considered for all actions, so a pruned reward function can be re-considered later. This may have the negative effect that all but the true one are pruned. In implementation, a minimum number of policies can be specified.

\begin{figure}[!th]
\centering
\begin{subfigure}[b]{0.23\textwidth}
\centering
	\includegraphics[scale=0.35]{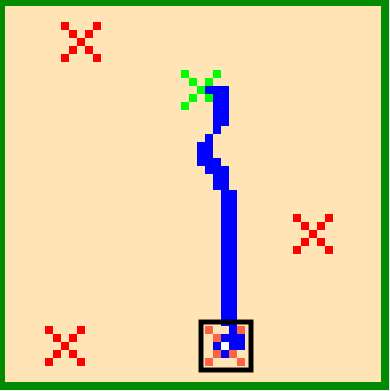}
	\caption{Without pruning}
	\label{fig:ambiguity-model-example:no-prune-empty}
\end{subfigure}
\begin{subfigure}[b]{0.23\textwidth}
\centering 
	\includegraphics[scale=0.35]{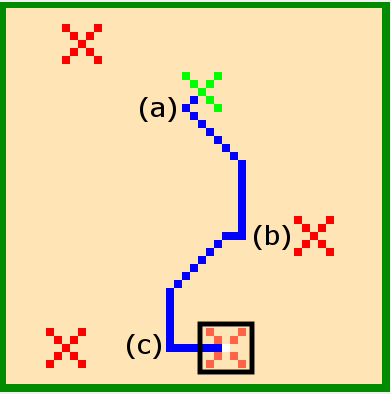}
	\caption{With pruning}
	\label{fig:ambiguity-model-example:prune-empty}
\end{subfigure}
\hfill
\caption{Examples of the ambiguity model. The agent navigates from the green starting point to the real destination (orange \& marked), using bogus destinations (red).}
\label{fig:ambiguity-model-example}
\end{figure}

Figure~\ref{fig:ambiguity-model-example} illustrates a path planning problem in which the agent must navigate from the green start point to the orange destination. The bogus destinations are red. In  Figure~\ref{fig:ambiguity-model-example:no-prune-empty}, the agent minimises information gain for all goals without pruning. It is difficult to see, but the thicker line  in Figure~\ref{fig:ambiguity-model-example:no-prune-empty} compared to Figure~\ref{fig:ambiguity-model-example:prune-empty} is the agent zigzagging repeatedly left-to-right. In Figure~\ref{fig:ambiguity-model-example:prune-empty} with pruning, at the first turn, labelled (a), the destination at the top left is pruned, while at turn (b), the destination on the right is pruned, and turn (c) prunes the destination at the bottom left. This delivers a shorter path than in Figure~\ref{fig:ambiguity-model-example:no-prune-empty} because it avoids zigzagging behaviour from trying to maximise the entropy of all destinations.

\subsection{Irrationality Model}

The irrationality model is based on \citet{masters2019goal}. The \emph{deceptive Q-value} of an action is a weighted sum of its optimal Q-value and a \emph{irrationality measure}. The higher the weight on the optimal Q-value, the less deceptive the behaviour.

First, we define the \emph{irrationality measure} for an observation sequence, which is dependent on the history of a sequence of actions, rather than a single action. This is because an action may appear rational in a one state, but not in the context of a longer sequence.

\begin{definition}[Irrationality Measure]
For an observed sequence of state-action pairs $\Vec{o}$, the \emph{irrationality measure} of $\Vec{o}$ with respect to reward function $r_i$ is:
\begin{equation}
  IM(\Vec{o}) = 1-\max_{r_i \in \mathcal{R}} 
  \Delta_{r_i}(\Vec{o})
 \end{equation}
in which $\Delta_{r_i}$ is a divergence function (Equation~\ref{eq:divergence-diff}).
This definition is similar to the definition of rationality for path planning outlined by \citet{masters2019goal}.
\end{definition}

Under this definition, a sequence $\Vec{o}$ that has a low value for \emph{all} reward functions has a high $IM$ --- it is irrational not to make progress towards at least one goal. We take the minimum of all  reward functions: if the sequence is rational for \emph{any} of the possible reward functions, then it is deemed rational by an observer who does not know the true reward function.

The goal of the agent is to maximise its expected reward as well as its irrationality. We use a parameter $\alpha$ $(0 \leq \alpha \leq 1)$ as the weight to define the importance of the Q-value versus the irrationality. The deceptive policy $\pi^D$ is defined as the weighed sum of the optimal Q-value and  the irrationality measure:
\begin{align}
    \pi^D(\Vec{o}, s)= \argmax_{a \in A}\ (1{-}\alpha) \underbrace{Q'_r(s,a)}_{\text{Optimal} } + \alpha \underbrace{IM(\Vec{o} \cdot (s,a))}_{\text{Irrational}}
\end{align}
in which $Q'_r(s,a)$ is $Q_r(s,a)$ normalised against other actions $a' \in A$ to range [0,1]. The higher  $\alpha$, the lower the weight given to the Q-value and the more irrational the behaviour. 

\begin{figure*}[!th]
\centering
\begin{subfigure}[b]{0.3\textwidth}
\centering
	\includegraphics[scale=0.25]{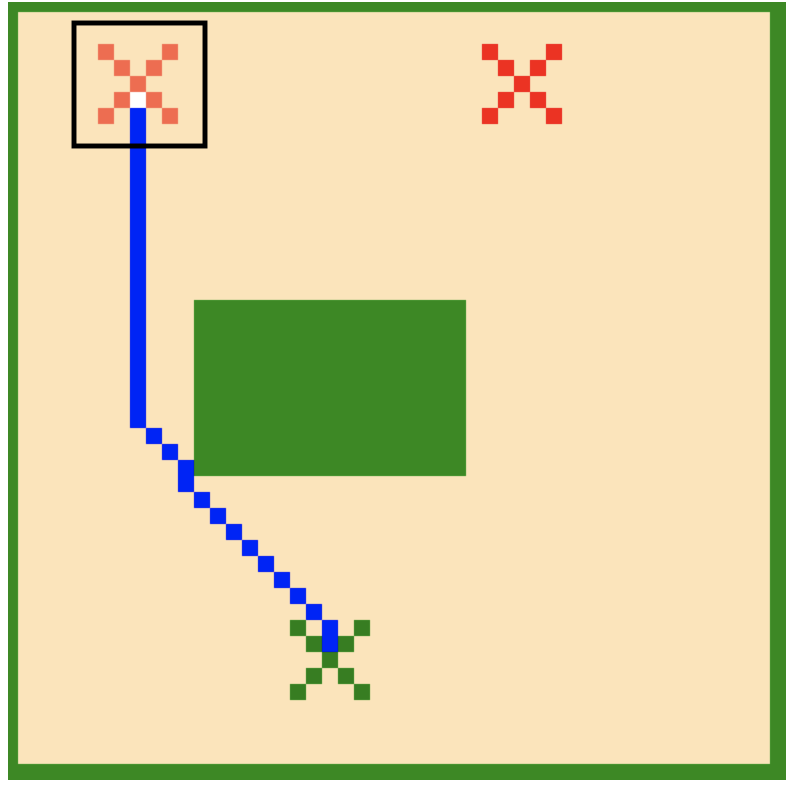}
	\caption{$\alpha=0$}
	\label{fig:irrationality-model-example:alpha-0}
\end{subfigure}
\begin{subfigure}[b]{0.3\textwidth}
\centering 
	\includegraphics[scale=0.25]{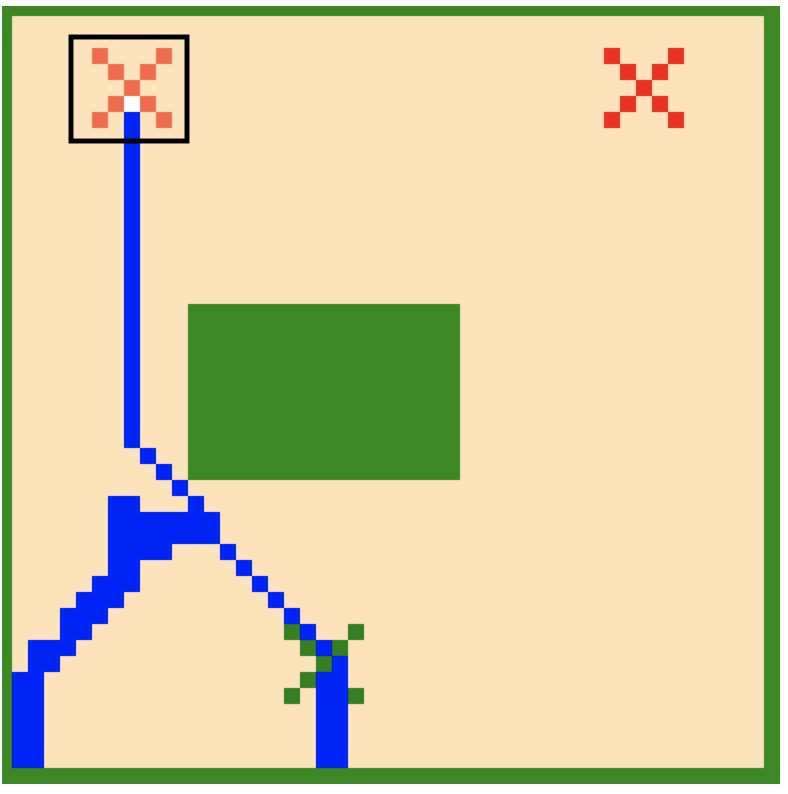}
	\caption{$\alpha=0.15$}
	\label{fig:irrationality-model-example:alpha-0-15}
\end{subfigure}
\begin{subfigure}[b]{0.3\textwidth}
\centering 
	\includegraphics[scale=0.25]{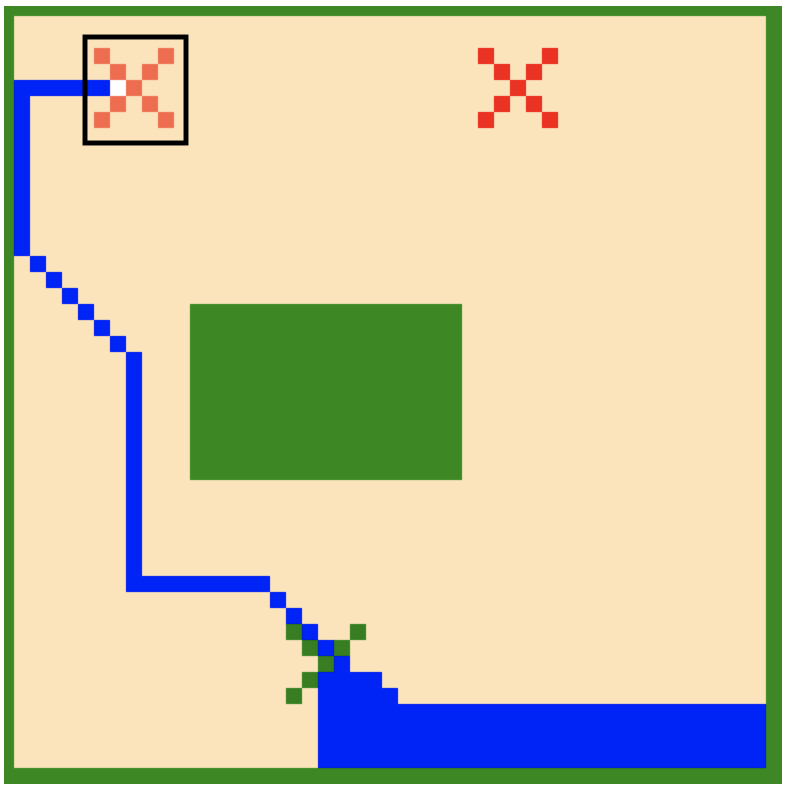}
	\caption{$\alpha=0.3$}
	\label{fig:irrationality-model-example:alpha-0-3}
\end{subfigure}
\caption{Examples of the irrationality model in path planning. The agent navigates from the green starting point to the real destination (orange \& marked), using bogus destinations (red).
}
\label{fig:irrationality-model-example}
\end{figure*}

Figure~\ref{fig:irrationality-model-example} illustrates the irrationality model in a path planning setting. When $\alpha=0$, we get honest behaviour. As $\alpha$ increases, rationality decreases. For $\alpha=0.15$ and $0.3$, the agent moves away from both destinations. For $\alpha=0.3$, the blue block in the bottom corner reflects the agent's excessive irrationality. As the agent moves towards its true destination, its behaviour becomes more rational, capturing a similar idea to the last deception point in \citet{masters2017deceptive}: it becomes more difficult to deceive as one `approaches' a goal. 

%% file: 4-computational-evaluation.tex
\section{Computational Evaluation}
\label{sec:computational-evaluation}

In this section, we present a computational evaluation our two models from Section~\ref{sec:model}. The goals of the experiment are to quantify how deceptive our models are compared to an honest baseline, and to determine the `cost' of deception based on how far the deceptive behaviour deviates from the honest optimal model.

\subsection{Experiment Design}

We implemented the ambiguity and irrationality models defined in Section~\ref{sec:model} and applied in two domains: the P4 path planning framework\footnote{See \url{https://bitbucket.org/ssardina-research/p4-simulator/}} and the Berkeley Pacman framework \footnote{See \url{http://ai.berkeley.edu/}}.

\textbf{Independent variables}~ We evaluated four different agents in our experiments:
\begin{enumerate}
    \item an `honest' agent, which uses the optimal policy for the real reward function;
    \item one deceptive agent that uses the ambiguity models; and
    \item[(3-4)] two deceptive agents using the irrationality model, with  $\alpha=0.3$ and $\alpha=0.5$ respectively.
\end{enumerate}

Reward functions gave a reward of 10,000 for reaching the destination state. Negative rewards model actions costs, with $-1$ for an action up, down, left, or right, and $-\sqrt{2}$ for any diagonal move. 
For the ambiguity model, we set pruning parameter $\delta=0$. Q-functions were implemented as Q-tables for P4 and were learnt using value iteration \cite{sutton2018reinforcement} until convergence, and implemented using linear function approximation for Pacman and were learnt using Q-learning.

\textbf{Measures}~ We measured: (1) the total path cost, which is the inverse of the discounted reward; (2) the probability assigned to the true reward function, calculated using a na\"ive intention recognition algorithm \cite{masters2017cost,vered16}; and (3) the \emph{simulation} value of the paths from \citeauthor{masters2017deceptive} \cite{masters2017deceptive}:
\[
 simulation(\vec{o_t}, \mathcal{R}) = \frac{\sum_{j=1}^t \cdot \max_{r_i \in \mathcal{R}\setminus \{r\}} P(r_i \mid \vec{o_t}) - P(r \mid \vec{o_t})}{t}
\]
This definition calculates, for each state-action pair, the deceptiveness of the step by using the  observer's intention recognition model and taking the difference between the most likely reward function and the actual reward function. The deceptiveness $D(\vec{o_t}, \mathcal{R})$ is then the average deceptiveness over the observed trace

This intention recognition algorithm used for measures 2 and 3 assigns the probability of each destination as being the real destination by comparing how rational the observed behaviour is for each destination, where rationality is assigned based on how close the observed behaviour is to the optimal behaviour.

\textbf{Experiment parameters}~ We used five different layouts for each domain, varying in size and structure. For example, for P4 we varied number and density of obstacles as follows:
\begin{enumerate}
    \item $49 \times 49$ with no obstacles, such as in Figure~\ref{fig:ambiguity-model-example};
    \item $49 \times 49$ with some large obstacles, such as in Figure~\ref{fig:irrationality-model-example};
    \item $49 \times 49$ map with random and high density obstacles;
    \item $100 \times 100$ with `archipelago' (a small number of large island obstacles);
    \item $100 \times 100$ with many rooms and corridors.
\end{enumerate}
For each layout type, we defined eight different variations by changing the number of goals (three or five in P4), distribution of rewards, and the position of the real reward, leading to a total of 40 layouts. For the Pacman domain, we used 10 maps from the Berkley framework.
Each model was applied to all 50 maps for each domain. For each path generated, the intention recognition measure was taken at nine `checkpoints': every $10\%$ along the path.

\subsection{Results}

\begin{figure*}[!ht]
\centering
    \includegraphics[scale=0.42]{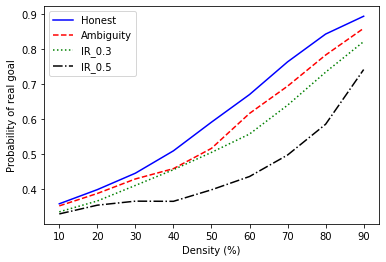}
    ~~
	\includegraphics[scale=0.42]{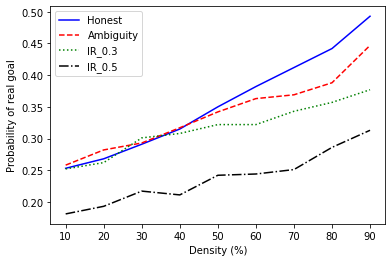}
	~~
	\includegraphics[scale=0.42]{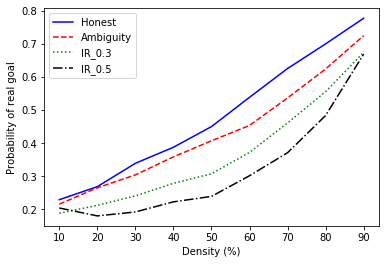}
	\caption{Intention recognition for P4 with three goals (left), P4 with 5 goals (middle) and Pacman (right)}
	\label{fig:computational-results:probability}
\end{figure*}

Figure~\ref{fig:computational-results:probability}  plots the average probability of the real `goal' at each point for both domains. We separate the results into three-goal maps in P4, five-goal maps in P4, and the Pacman maps.
In all cases, it is easier to identify the real reward function as more plan steps are revealed, however, there is a clear trend that the deceptive models make it more difficult. The irrationality models are more deceptive but this must be considered alongside path costs. 

\begin{figure*}[!ht]
\centering
    \includegraphics[scale=0.42]{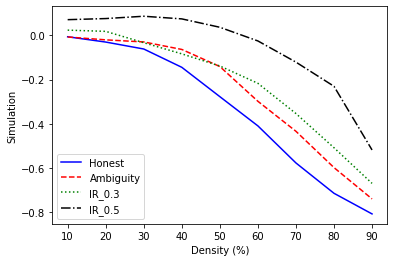}
    ~~
	\includegraphics[scale=0.42]{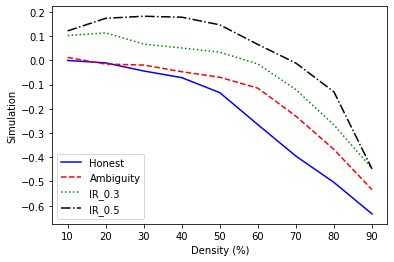}
	~~
	\includegraphics[scale=0.42]{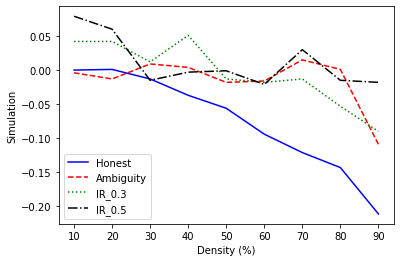}
	\caption{Intention recognition for P4 with three goals (left), P4 with 5 goals (middle) and Pacman (right)}
	\label{fig:computational-results:simulation}
\end{figure*}

\begin{figure*}[!ht]
\centering
	\includegraphics[scale=0.32]{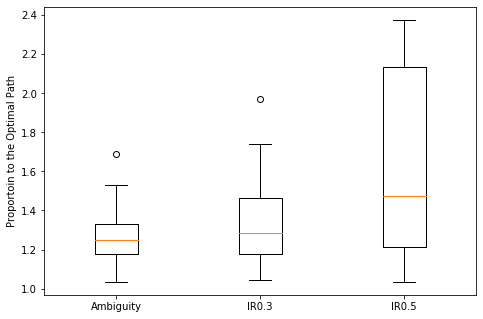}
	~~
	\includegraphics[scale=0.32]{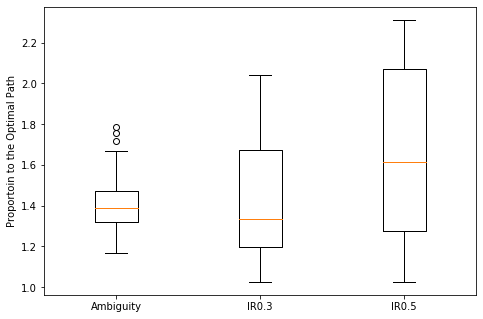}
	~~
	\includegraphics[scale=0.32]{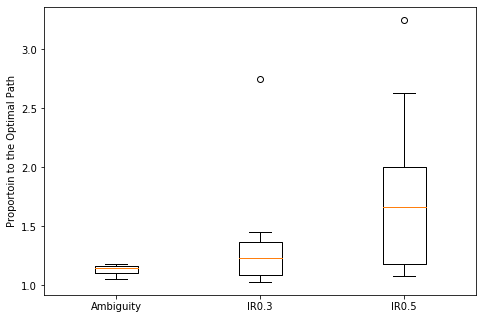}
\caption{Path costs, proportional to honest path for P4 with three goals (left), P4 with 5 goals (middle) and Pacman (right)}
\label{fig:computational-results:path-costs}
\end{figure*}

Figure~\ref{fig:computational-results:simulation} demonstrates the simulation measurements for different scenarios. The simulation level of the honest model is the lowest among all models in most of the time, which consists with the probability results. This is perhaps the most interesting measure in our experiment, as it measures true `deceptiveness'. 

On the simulation data, we performed a Kolmogorov-Smirnov test of normality to confirm that our data matches the characteristics of a normal distribution. We then performed paired t-tests for independent samples between the honest model and the three deceptive models.
The honest model showed the lowest level of simulation ($M$=-0.24, $SD$=0.17). In comparison, the ambiguity model ($M$=-0.16, $SD$=0.18) was more deceptive than the honest model  $t(98)$=2.05, $p$=.04. Similarly, the IR\_0.3 model ($M$=-0.13,$SD$=0.24) was more deceptive than the honest model $t(98)$=2.64, p=.009, as was the IR\_0.6 model ($M$=-0.009,$SD$=0.21), $t(98)$=6.06, $p<.001$.

Figure~\ref{fig:computational-results:path-costs} shows  the path costs as a proportion of the length of optimal (honest) path. The ambiguity model arrives at the destination with fewer actions than either irrationality model. This is important because in addition to being deceptive, the objective of deceptive reinforcement learning is to maximise discounted expected rewards. In some cases, irrationality model with $\alpha$=$0.5$ was more than four times as long. If we give higher priority to the expected reward for the real reward function, we may prefer the ambiguity model or to use the irrationality model with a lower value of $\alpha$. In some cases, if deception is weighted low enough, the honest model would still be preferred because of the short paths, which results in higher actual reward for discount factor $\gamma < 1$.

Analysis looking into individual maps, we see that the IR models generate longer paths due to the randomness in the paths. For the ambiguity model, the paths are slightly longer for the five-goal maps, because there are more bogus goals that `pull' the agent away from the optimal honest path. This indicates that, in some domains in which the real reward is strongly weighted, even if there are many possible bogus reward functions, it may still be more suitable to select only a subset of the bogus goals for the entropy calculations. Further results on individual maps are available in the supplementary material.

Overall, we see that it is easier to deceive in the Pacman game than in path planning, which we attribute to the fact that there is just a single reward at the destination in path planning, and eventually we end up with fewer and fewer goals until finally the only likely goal from the observer's view is the real goal.

%% file: 5-human-evaluation.tex
\section{Human Behavioural Evaluation}
\label{sec:human-evaluation}

In this section, we describe a human behavioural experiment to measure the ability to deceive people, rather than algorithms. Participants were aware that they could be deceived, unlike the `na\"ive' intention recognition model.
There is only one intention recognition model that detects deception for sequential decision-making: the irrationality model by \citet{masters2019goal}. However, as our irrationality model uses this concept to generate behaviour, the \citeauthor{masters2019goal} model is not valid for us.

\subsection{Experiment Design}

The experiment design was similar to that used for the computational evaluation, with three exceptions: (1) instead of the intention recognition algorithm, we ask human participants to estimate the goal distribution; (2) the human participants were provided with only a random selection of the maps and methods; and (3) we assessed based only on the path-planning problem. 

Our experiment used 40 maps $\times$ four possible models producing behaviour = 160
map-path pairs. We generated checkpoints at 25\%, 50\%, and 75\%, leading to 480 stimuli in total. Each participant received 38 randomly-selected stimuli.
Participants were `aware'; that is, they were explicitly told  that the agent may  try to hide its true destination, and that they should try to guess the true destination.
We recruited 69 participants via Amazon Mechanical Turk, a crowd-sourcing platform often used  for human-subject experiments~\cite{buhrmester2018evaluation}. Participants were paid US\$4 for completing all tasks, which took on average 11.5 minutes. Participants were aged 20-55 ($\mu = 32)$. 15 participants were female,  54 were male, and none chose to specify their gender manually.

\subsection{Results}

\begin{figure*}[!t]
\centering
\begin{subfigure}[b]{0.45\textwidth}
\centering 
	\includegraphics[scale=0.45]{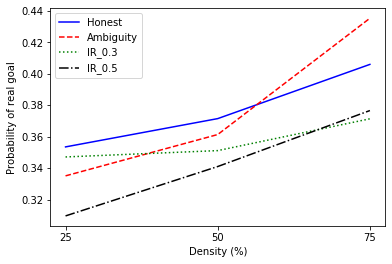}
	\caption{Three goals}
	\label{fig:human-results:3-goals}
\end{subfigure}
\begin{subfigure}[b]{0.45\textwidth}
\centering 
	\includegraphics[scale=0.45]{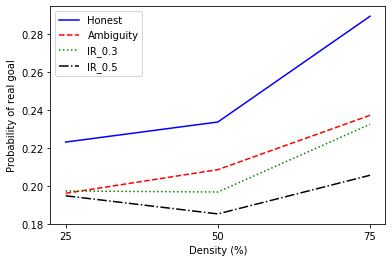}
	\caption{Five goals}
	\label{fig:human-results:5-goals}
\end{subfigure}
\caption{Average of experiment participant's prediction of the true destination for the two scenarios}
\label{fig:human-evaluation-results}
\end{figure*}

\begin{figure*}[!t]
\centering
\begin{subfigure}[b]{0.45\textwidth}
\centering 
	\includegraphics[scale=0.45]{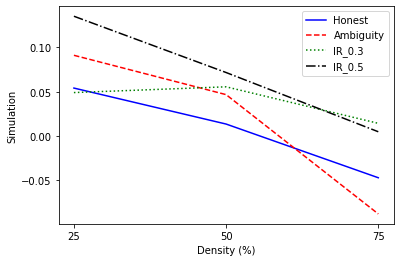}
	\caption{Three goals}
	\label{fig:human-results:simulation-3-goals}
\end{subfigure}
\begin{subfigure}[b]{0.45\textwidth}
\centering 
	\includegraphics[scale=0.45]{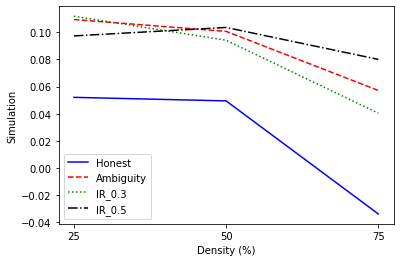}
	\caption{Five goals}
	\label{fig:human-results:simulation-5-goals}
\end{subfigure}
\caption{Average simulation based on the experiment participants' prediction for the two scenarios}
\label{fig:human-evaluation-simulation-results}
\end{figure*}

Figures~\ref{fig:human-evaluation-results} and \ref{fig:human-evaluation-simulation-results} summarise the results for the human subject evaluation. We see similar outcomes to that of the na\"ive intention recognition algorithm, except that human subjects were overall less accurate than the na\"ive model, even for honest behaviour. This is understandable as the optimal behaviour is straightforward for an algorithm to calculate, but less so for a human. At the first checkpoint, by which point participants have seen 25\% of the path, the accuracy is close to random.

For the ambiguity model with three goals, participants were more accurate than for the honest model at 75\% density, but this is mostly accounted for by noisy data -- the difference is less than 3\%. 
The deceptive models were more effective at deceiving in the five-goal model than the three-goal model, which is unsurprising as there are more bogus goals to use.

An interesting point is the effect of the participants being aware that they are being deceived, which is not the case for the earlier computational experiments in which the observer model is na\"ive. In the computational experiments, the honest model is never considered deceptive. The simulation value is at most 0, meaning that the real destination is judged to be as likely as others. However, in the human subject experiments, the honest model is, on average, considered to be deceptive early in the experiment, presumably because the participants were assuming that the model was using deception as simulation (showing the false). Also interesting is that the deceptive models were considered deceptive right up until the 75\% mark and presumably beyond. In the computational experiments, the simulation value was, on average, below 0 at the 75\% mark for all deceptive models. This is perhaps due to the fact that the human participants are unable to make accurate judgements as quickly as the intention recognition algorithm. As such, results may differ if we had a non-na\"ive intention recognition model.

Overall, we see that our models deceived participants for the path planning task, but the effectiveness may not be sufficient if the length of the plan is considered too high. This largely depends on the weight $\omega$ in Equation~\ref{eq:deceptive-L}.

\subsection{Limitations}

There are several limitations with our study.
First, while path planning is a good application for human behavioural experiments (people are good at spatial reasoning), it is only one domain, so further experimentation on different types of domains is necessary. Second, the na\"{i}ve intention recognition model we used to evaluate deception in the computational evaluation is not as sophisticated as our model of deception -- it does not mitigate for the fact that it is being deceived. This is difficult to mitigate because we need a level of separation between the methods and the evaluation metrics, and the only suitable model of which we are aware is the irrationality model \cite{masters2019goal}, on which our model is based. Third, there was only minimal incentive for our experimental participants, which is not reflective of some applications where failing to identify deception can have devastating outcomes.

%% file: 6-conclusions.tex
\section{Discussion and Future Work}
\label{sec:conc}

In this paper, we presented two models for preserving the privacy of reward functions in reinforcement learning. 
Through computational and human evaluation in a path planning task, we have shown that the models can deceive both na\"ive intention recognition algorithms and human subjects. However, often the length of plans is significantly higher, meaning that for domains in which deception is weighted only lightly, an honest agent may be more suitable. Clearly, this judgement depends on the domain and the measure of deception used.

In future work, we will apply this model to more tasks, and we will investigate this model in policy-based reinforcement learning, in which we do not have Q-functions, but learn a policy directly. Further, we aim to extend these models to models similar to that in \citet{ornik2018deception}, in which the observer has only partial observability of the agent and the environment.

\emph{Acknowledgements} This paper was supported by ARC Grant DP180101215 \emph{A Computational Theory of Strategic Deception}.  Ethics approval was obtained from The University of Melbourne Human Ethics Committee; ethics approval number 1954358.1.

%% file: appendix.tex
\appendix 
\section{Supplementary Material: Non-deceptive actions}

This figure shows the percentage of non-deceptive actions, as measured by whether the observer is deceived after the execution of the action, for each of the 50 cases in the computational experiments.

\begin{center}
    \includegraphics[angle=270,scale=0.65]{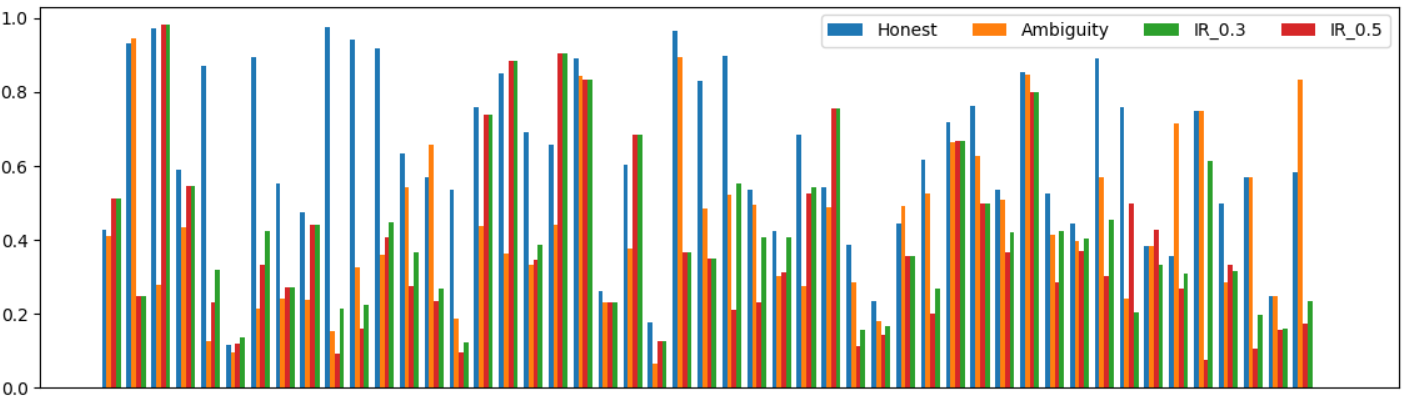}
\end{center}